\documentclass[10pt,twocolumn,letterpaper]{article}

\usepackage{iccv}
\usepackage{times}
\usepackage{epsfig}
\usepackage{graphicx}
\usepackage{amsmath}
\usepackage{amssymb}
\usepackage{booktabs}
\usepackage{multirow}

\usepackage[pagebackref=true,breaklinks=true,letterpaper=true,colorlinks,bookmarks=false]{hyperref}

\iccvfinalcopy 

\def\iccvPaperID{5205} 

\ificcvfinal\fi

\begin{document}

\title{Body Knowledge and Uncertainty Modeling for Monocular 3D Human Body Reconstruction}

\author{Yufei Zhang$^1$, Hanjing Wang$^1$, Jeffrey O. Kephart$^2$, Qiang Ji$^1$ \\
$^1$Rensselaer Polytechnic Institute, $^2$IBM Research \\
{\tt\small \{zhangy76, wangh36, jiq\}@rpi.edu, kephart@us.ibm.com}
}

\maketitle
\ificcvfinal\thispagestyle{empty}\fi

\begin{abstract}
   
While 3D body reconstruction methods have made remarkable progress recently, it remains difficult to acquire the sufficiently accurate and numerous 3D supervisions required for training. In this paper, we propose \textbf{KNOWN}, a framework that effectively utilizes body \textbf{KNOW}ledge and u\textbf{N}certainty modeling to compensate for insufficient 3D supervisions. KNOWN exploits a comprehensive set of generic body constraints derived from well-established body knowledge. These generic constraints precisely and explicitly characterize the reconstruction plausibility and enable 3D reconstruction models to be trained without any 3D data. Moreover, existing methods typically use images from multiple datasets during training, which can result in data noise (\textit{e.g.}, inconsistent joint annotation) and data imbalance (\textit{e.g.}, minority images representing unusual poses or captured from challenging camera views). KNOWN solves these problems through a novel probabilistic framework that models both aleatoric and epistemic uncertainty. Aleatoric uncertainty is encoded in a robust Negative Log-Likelihood (NLL) training loss, while epistemic uncertainty is used to guide model refinement. Experiments demonstrate that KNOWN's body reconstruction outperforms prior weakly-supervised approaches, particularly on the challenging minority images. 
\end{abstract}

\section{Introduction}
\label{sec:intro}

Recovering 3D human body configurations from monocular RGB images has broad applications in robotics, human-computer interaction, and human behaviour analysis. This is a challenging task, as the inherent depth ambiguity under-constrains the reconstruction problem. To alleviate the high dimensionality of the reconstruction space, deformable body models such as SMPL~\cite{loper2015smpl} or GHUM~\cite{xu2020ghum} have been used widely to represent a 3D human body in terms of body pose and shape parameters. Building upon these parametric models with the aid of deep learning, model-based methods have made promising progress on existing benchmarks~\cite{pavlakos2018learning,kanazawa2018end,kolotouros2019learning,jiang2020coherent,choi2020pose2mesh,li2021hybrik,lin2021end,kolotouros2021probabilistic,kocabas2021pare}. However, obtaining 3D annotations for training deep models often requires a Motion Capture (MoCap) system~\cite{vonMarcard2018}, which is cumbersome, expensive, and limited to specific environments and subjects. Moreover, labeling noise can be introduced when obtaining the 3D annotations by fitting parametric models to sparse 3D MoCap markers~\cite{loper2014mosh,mahmood2019amass} or generating pseudo-labels using existing reconstruction models~\cite{joo2021exemplar,moon2022neuralannot,li2022cliff}. Thus it is important to reduce the dependency on 3D annotations to the greatest possible extent.

One approach
is to employ weak supervisions, such as utilizing 2D landmark annotations combined with prior knowledge of constraints on human body pose and shape that are extracted from data~\cite{bogo2016keep,kanazawa2018end,pavlakos2019expressive,hassan2019resolving,zanfir2018monocular,zhou2017towards,dabral2018learning,habibie2019wild,yang20183d}. While this method avoids the demand for 3D annotations, this problem is replaced by another: it is challenging to collect a sufficiently homogeneous and dense data set from which to extract a precise prior~\cite{herda2005hierarchical}. Moreover, such data-driven priors do not explicitly capture reconstruction plausibility, as they are only modeled via a black-box distribution. In this work, we avoid data-driven priors by exploiting generic body constraints derived from well-established studies on human body structure and movements~\cite{hamill2006biomechanical,nasa1995std,winter2009biomechanics} to effectively utilize
2D annotations for 3D body reconstruction.

Another problematic aspect of existing approaches is that they typically combine images from multiple datasets for training, neglecting the subsequent problem of data noise and data imbalance. In particular, data noise can stem from inconsistent joint definitions across different datasets~\cite{sarandi2023learning} and the presence of low quality images~\cite{kocabas2021pare,xu20213d}. Data imbalance occurs because 3D datasets are rich in images from videos but suffer from diversity in subjects and actions, whereas 2D datasets are diverse in subjects, poses, and backgrounds but usually suffer from having few images per subject-pose-background combination. KNOWN deals with the problems of data noise and data imbalance by modeling uncertainty. Specifically, KNOWN employs a novel probabilistic framework that captures both aleatoric uncertainty (also called data uncertainty), which measures the inherent data noise, and epistemic uncertainty (also called model uncertainty), which reflects the lack of knowledge due to limited data~\cite{kendall2017uncertainties,Lakshminarayanan_NIPS17_ensemble,striking2019CVPR}. KNOWN is trained using NLL, which achieves robustness to data noise by adaptively assigning weights based on the captured uncertainty. Because previous methods treat all the data samples equally, their performance suffers on the minority images (those that are distinct from the training data and have low data density). KNOWN uses an uncertainty-guided refinement strategy to improve model performance, particularly on these minorities.

In summary, our main contributions include: 
\begin{itemize}
    \item A systematic study of body knowledge and its encoding as generic constraints that allow training 3D body reconstruction models without requiring any 3D data;
    \item The first 3D body reconstruction model that accounts for both aleatoric and epistemic uncertainty, thereby supporting both data noise and data density characterization and model performance improvement;
    \item Extensive experiments that demonstrates improved reconstruction accuracy over existing works. In particular, KNOWN outperforms fully-supervised 3D reconstruction models on challenging minority test images. 
\end{itemize}


\section{Related Work}

\subsection{3D Body Reconstruction with Prior Knowledge}

Prior knowledge can be extracted from data (data-driven prior) or derived from established body knowledge (generic body prior). Here we review how the existing works encode and leverage these two types of priors.

\textbf{Data-driven priors} are typically learned from MoCap data~\cite{cmumocap,akhter2015pose,mahmood2019amass}. Some optimization-based methods encode pose feasibility via nonlinear inequality constraints~\cite{akhter2015pose}, a Gaussian Mixture Model~\cite{bogo2016keep,arnab2019exploiting}, or a Variational Auto-Encoder~\cite{pavlakos2019expressive}. They incorporate these learned priors as constraints or penalties to avoid infeasible joint angle estimation. Additionally, average bone length of training data is exploited to ensure body scale validity ~\cite{benabdelkader2008statistical,zhou20153d,ramakrishna2012reconstructing,wang2014robust}. Among learning-based methods, adversarial framework~\cite{jack2017adversarially,yang20183d,wandt2019repnet,chen2019unsupervised,kundu2020appearance,zhang2020inference,gong2021poseaug,yu2021towards} or normalizing flow~\cite{zanfir2020weakly,biggs20203d,wehrbein2021probabilistic,kolotouros2021probabilistic,zanfir2021neural,zanfir2021thundr} is employed to encode body pose feasibility. These models are utilized in building prior body models for constrained parameter estimation or to regularize model training. In particular, HMR \cite{kanazawa2018end} learns a pose and shape prior from MoCap data, which facilitates subsequent training on images that possess only 2D keypoint annotations. These data-driven priors, however, assume that sufficient 3D data be available. In contrast, KNOWN does not require any 3D data or annotations.

\textbf{Generic body priors} recognize that human body structure and pose comply with basic functionality principles that generalize across different people and activities~\cite{hamill2006biomechanical,nasa1995std,winter2009biomechanics}. Some works design a neural network architecture to encode body knowledge  \cite{lee2018propagating,fang2018learning,ci2019optimizing,guler2019holopose,choi2020pose2mesh,chen2020towards,zeng2020srnet,xu2020deep,yu2021skeleton2mesh,li2021hybrik}, such as predicting joint angles from the torso outward based on body kinematics~\cite{georgakis2020hierarchical}. These models still require either 3D supervisions or pose priors for training. Moreover, some authors~\cite{wei2009modeling,dabral2018learning} introduce bone symmetry loss to reduce the dependency on 3D annotations. We supplement this body anatomy constraint by adding statistics of different bones from an anthropometric study~\cite{hamill2006biomechanical} and by introducing special geometry characteristics of human body joints. Other authors~\cite{chen2013data,habermann2020deepcap,spurr2020weakly} utilize body biomechanics by imposing constraints on individual joint rotations. We supplement these constraints by adding inter-joint dependencies based on body functional anatomy \cite{hamill2006biomechanical}. Additionally, Proxy Geometries~\cite{bogo2016keep,zanfir2018monocular} and Signed Distance Field (SDF)~\cite{pavlakos2019expressive,Karras:2012:MPC:2383795.2383801,Tzionas:IJCV:2016} are used in optimization-based frameworks to guard against a key body physics constraint: body parts should not inter-penetrate. We introduce a SDF-based loss~\cite{pavlakos2019expressive} into our learning framework to avoid collisions. To summarize, existing works consider body knowledge partially, while we combine constraints comprehensively based upon body anatomy, biomechanics, and physics and properly encode them into a probabilistic model that produces valid 3D predictions without requiring any 3D data.

\subsection{3D Body Reconstruction with Uncertainty}

Although uncertainty modeling has shown benefits in various computer vision tasks ~\cite{striking2019CVPR,wang2021data,kendall2017uncertainties}, such as robust 3D face model fitting via incorporating 2D landmark prediction uncertainty~\cite{wood20223d}, it has not been well explored in 3D body reconstruction. For 3D body reconstruction, some methods output a probability distribution to generate multiple hypotheses to account for the depth ambiguity \cite{kolotouros2021probabilistic,wehrbein2021probabilistic,biggs20203d}. However, they do not explicitly quantify uncertainty, nor do they resolve it into aleatoric and epistemic components. Other methods quantify aleatoric uncertainty and relate it with image occlusion \cite{sengupta2021probabilistic,sengupta2021hierarchical}. We are unaware of any existing works that quantify epistemic uncertainty. Furthermore, there is a noticeable lack of research on effectively leveraging uncertainty for model improvements.

To our knowledge, KNOWN is the first 3D body reconstruction model that captures both aleatoric and epistemic uncertainties. It does so using a single two-stage probabilistic neural network, the uncertainty quantification efficiency of which compares favorably with standard Bayesian~\cite{geman1984stochastic,SGHMC_chen2014,hastings1970monte,blei2017variational} and non-Bayesian~\cite{Lakshminarayanan_NIPS17_ensemble,wen2020batchensemble,gal2016dropout,folgoc2021mc,van2020uncertainty} methods. 
Moreover, KNOWN is unique in that it leverages epistemic uncertainty to improve reconstruction accuracy and (unlike existing probabilistic 3D body reconstruction models) it does not require 3D data.


\begin{figure*}[t]
\vspace{-0.3cm}
\begin{center}
\includegraphics[width=0.98\linewidth]{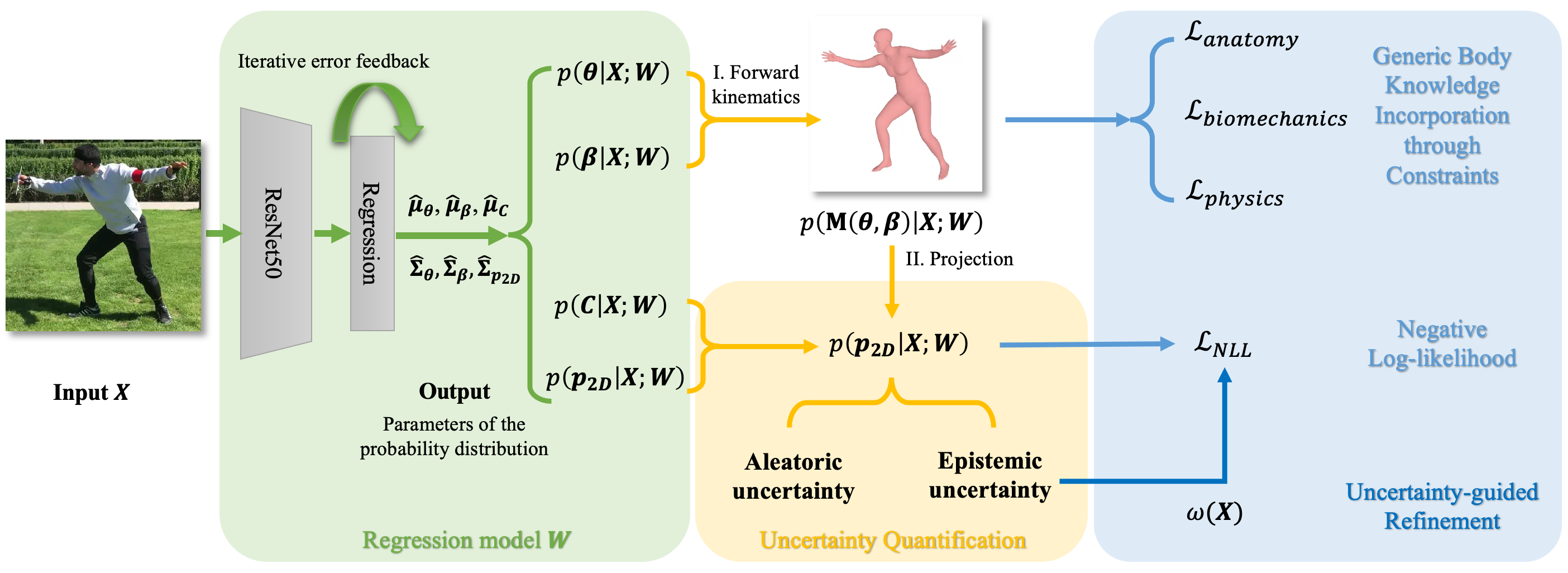}
\end{center}
\vspace{-0.2cm}
\caption{\textbf{Method overview.} KNOWN employs a two-stage probabilistic neural network for 3D body reconstruction from 2D images. Given an input image, the first stage models the conditional probability distribution of the body pose, shape, and camera parameters, while the second stage captures the conditional distribution of the corresponding image projection.
}
\label{fig:overview}
\vspace{-.2cm}
\end{figure*}

\section{Method}

As illustrated in Fig.~\ref{fig:overview}, KNOWN is a two-stage probabilistic neural network. As will be elaborated in Sec.~\ref{sec:twostageprobablisticmodel}, KNOWN captures both aleatoric and epistemic uncertainty. It is trained by combining the generic body constraints (Sec.~\ref{sec:GBC}) with a new NLL loss (Sec.~\ref{sec:NLL}). KNOWN's final step (uncertainty-guided refinement) and the total training loss are described in Sec.~\ref{sec:UncertaintyGuided}.

To represent a 3D human body, we use SMPL \cite{loper2015smpl}, which includes the pose parameters $\boldsymbol{\theta} \in \mathbb{R}^{23\times6}$ and the shape parameters $\boldsymbol{\beta}\in \mathbb{R}^{10}$ to characterize the rotation of 23 body joints and body shape variation, respectively. To establish 3D-2D consistency, we employ a weak-perspective projection model with parameters $\mathbf{C}=[{s,\mathbf{R},\mathbf{t}}]$, where $\mathbf{t}\in\mathbb{R}^{2}$, $s \in \mathbb{R}$, and $\mathbf{R}\in\mathbb{R}^{6}$ denotes the global translation, scale factor, and camera rotation, respectively. The projection of 3D body keypoints is $\mathbf{p}_{2D}=Proj(\mathbf{P}_{3D},\boldsymbol{\beta},\mathbf{C})$, where $Proj(\cdot)$ denotes the weak-perspective projection function. We use a 6D representation for the body pose and camera rotation to avoid the singularity problem \cite{zhou2019continuity}.

\subsection{Two-stage Probabilistic Regression Model}
\label{sec:twostageprobablisticmodel}

As shown in Fig.~\ref{fig:overview}, Stage I models the distributions of the body model and camera parameters given an input image, while stage II models the distributions of the corresponding 2D projections of body keypoints. As Gaussian distributions have been widely used to efficiently and effectively model continuous data, we assume Gaussian parameters. To specify the modeled probability distributions, we employ a ResNet50~\cite{he2016identity} as a backbone to extract image features plus a regression network with iterative error feedback~\cite{carreira2016human} to predict the distribution parameters. Details of the modeled distributions are introduced below.

\noindent \textbf{Stage I:} The distributions of the body model and camera parameters dependent upon an input image $\mathbf{X}$ and neural network weights $\mathbf{W}$ are:
\begin{align}
    p(\boldsymbol{\theta}|\mathbf{X};\mathbf{W})&= \mathcal{N}\big(\boldsymbol{\mu}_{\boldsymbol{\theta}}(\mathbf{X};\mathbf{W}),\boldsymbol{\Sigma}_{\boldsymbol{\theta}}(\mathbf{X};\mathbf{W})\big), \label{eq:inferpose} \\
   p(\boldsymbol{\beta}|\mathbf{X};\mathbf{W})&= \mathcal{N}\big(\boldsymbol{\mu}_{\boldsymbol{\beta}}(\mathbf{X};\mathbf{W}),\boldsymbol{\Sigma}_{\boldsymbol{\beta}}(\mathbf{X};\mathbf{W})\big), \label{eq:infershape}\\
    p(\mathbf{C}|\mathbf{X};\mathbf{W})&= \mathcal{N}\big(\boldsymbol{\mu}_{\mathbf{C}}(\mathbf{X};\mathbf{W}),\boldsymbol{\Sigma}_{\mathbf{C}}(\mathbf{X};\mathbf{W})\big) \label{eq:infercam}, 
\end{align}
where $\boldsymbol{\mu}_{\boldsymbol{\theta, \beta, C}}$ and $\boldsymbol{\Sigma}_{\boldsymbol{\theta, \beta, C}}$ are mean and covariance matrices. A nonlinear function $\mathbf{M}(\boldsymbol{\theta},\boldsymbol{\beta})\in \mathbb{R}^{6890\times3}$ that represents a forward kinematic process can be applied to obtain the 3D vertex positions $\mathbf{M}$. Then the 3D body joint positions are computed as a linear combination of the vertex positions via $\mathbf{P}_{3D}(\boldsymbol{\theta},\boldsymbol{\beta})= \mathbf{H}\mathbf{M}(\boldsymbol{\theta},\boldsymbol{\beta})$, where $J$ denotes the number of body joints, and $\mathbf{H}\in \mathbb{R}^{J\times6890}$ is a predefined joint regressor.

\noindent \textbf{Stage II:} The distribution of the corresponding projection of 3D body keypoints is assumed to be
\begin{equation}
    \label{eq:computedatauncertainty}p(\mathbf{p}_{2D}|\mathbf{Y},\mathbf{C};\mathbf{X},\mathbf{W})= \mathcal{N}\big(\boldsymbol{\mu}_{\mathbf{p}_{2D}}(\mathbf{Y},\mathbf{C}),\boldsymbol{\Sigma}_{\mathbf{p}_{2D}}(\mathbf{X};\mathbf{W})\big),
\end{equation}
where $\mathbf{Y}=[\boldsymbol{\theta},\boldsymbol{\beta}]$, $\boldsymbol{\mu}_{\mathbf{p}_{2D}}(\mathbf{Y},\mathbf{C})$ is the mean that is computed from 3D body keypoints $\mathbf{P}_{3D}(\mathbf{Y})$ and camera parameters $\mathbf{C}$ using the projection function, and $\boldsymbol{\Sigma}_{\mathbf{p}_{2D}}(\mathbf{X};\mathbf{W})$ is the covariance matrix that is directly estimated by the neural network. Given the definition above, we obtain the conditional probability distribution of $\mathbf{p}_{2D}$ as 
\begin{equation}
    \resizebox{.9\hsize}{!}{$p(\mathbf{p}_{2D}|\mathbf{X};\mathbf{W})= \iint p(\mathbf{p}_{2D}|\mathbf{Y},\mathbf{C};\mathbf{X},\mathbf{W})p(\mathbf{Y},\mathbf{C}|\mathbf{X};\mathbf{W})d\mathbf{Y}d\mathbf{C}.
    $}
\end{equation}
Assuming that the body pose $\boldsymbol{\theta}$, shape $\boldsymbol{\beta}$, and camera parameters $\mathbf{C}$ are independent, $p(\mathbf{Y},\mathbf{C}|\mathbf{X};\mathbf{W}) =p(\boldsymbol{\theta}|\mathbf{X};\mathbf{W})p(\boldsymbol{\beta}|\mathbf{X};\mathbf{W})p(\mathbf{C}|\mathbf{X};\mathbf{W})$.

\textbf{Inference and uncertainty quantification:} During inference, the final 3D human body is recovered via the estimated mean pose and shape, which are the mode of the respective distributions (Eq.~(\ref{eq:inferpose}-\ref{eq:infershape})). Then, following the typical uncertainty quantification strategy~\cite{kendall2017uncertainties,valdenegro2022deeper}, we compute the covariance matrix of the keypoint projection to quantify the aleatoric and epistemic uncertainties:
\begin{equation}\label{eq:total_covariance}
\resizebox{.95\hsize}{!}{$
\begin{split}
  \underbrace{\operatorname{Cov}_{p(\mathbf{p}_{2D}|\mathbf{X};\mathbf{W})}[\mathbf{p}_{2D}]}_{\text{Total uncertainty}}
  &=\underbrace{\operatorname{E}_{p(\mathbf{Y},\mathbf{C}|\mathbf{X};\mathbf{W})}\big[\operatorname{Cov}_{p(\mathbf{p}_{2D}|\mathbf{Y},\mathbf{C};\mathbf{X},\mathbf{W})}[\mathbf{p}_{2D}]\big]}_{\text{Aleatoric uncertainty}}\\
  &+\underbrace{\operatorname{Cov}_{p(\mathbf{Y},\mathbf{C}|\mathbf{X};\mathbf{W})}\big[\operatorname{E}_{p(\mathbf{p}_{2D}|\mathbf{Y},\mathbf{C};\mathbf{X},\mathbf{W})}[\mathbf{p}_{2D}]\big]}_{\text{Epistemic uncertainty}}.
\end{split}$}
\end{equation}
Based on Eq.~(\ref{eq:computedatauncertainty}), the aleatoric uncertainty is
\begin{align}
\label{aleatoric_uncertainty}
 \begin{split}
 &\underbrace{\operatorname{E}_{p(\mathbf{Y},\mathbf{C}|\mathbf{X};\mathbf{W})}\big[\operatorname{Cov}_{p(\mathbf{p}_{2D}|\mathbf{Y},\mathbf{C};\mathbf{X},\mathbf{W})}[\mathbf{p}_{2D}]\big]}_{\text{Aleatoric uncertainty}} \\
 & = \operatorname{E}_{p(\mathbf{Y},\mathbf{C}|\mathbf{X};\mathbf{W})}\big[\boldsymbol{\Sigma}_{\mathbf{p}_{2D}}(\mathbf{X};\mathbf{W})\big] \\
 & = \boldsymbol{\Sigma}_{\mathbf{p}_{2D}}(\mathbf{X};\mathbf{W}).
 \end{split}
\end{align}
The epistemic uncertainty is
\begin{equation}
\label{epistemic_uncertainty}
    \begin{split}
   &\underbrace{\operatorname{Cov}_{p(\mathbf{Y},\mathbf{C}|\mathbf{X};\mathbf{W})}\big[\operatorname{E}_{p(\mathbf{p}_{2D}|\mathbf{Y},\mathbf{C};\mathbf{X},\mathbf{W})}[\mathbf{p}_{2D}]\big]}_{\text{Epistemic uncertainty}} \\
   &= \operatorname{Cov}_{p(\mathbf{Y},\mathbf{C}|\mathbf{X};\mathbf{W})}\big[ \boldsymbol{\mu}_{\mathbf{p}_{2D}}(\mathbf{Y},\mathbf{C})\big].
    \end{split}
\end{equation}
The aleatoric uncertainty equals to the predicted variance as derived in Eq.~\eqref{aleatoric_uncertainty}, while computing the epistemic uncertainty in Eq.~\eqref{epistemic_uncertainty} is difficult because $\boldsymbol{\mu}_{\mathbf{p}_{2D}}(\mathbf{Y},\mathbf{C})$ is a nonlinear function of $\mathbf{Y}$ and $\mathbf{C}$ (the forward kinematic and projection functions are nonlinear). We hence use samples to approximate the covariance matrix. Denote $\{\mathbf{Y}^s\}_{s=1}^S $ and $\{\mathbf{C}^s\}_{s=1}^S $ as the samples drawn from $p(\mathbf{Y}|\mathbf{X};\mathbf{W})$ and $p(\mathbf{C}|\mathbf{X};\mathbf{W})$, respectively. We compute the keypoint projection of each sample and obtain $\{\boldsymbol{\mu}_{\mathbf{p}_{2D}}^s\}_{s=1}^S$. The epistemic uncertainty is computed as
\begin{equation}
\begin{split}
 &\underbrace{\operatorname{Cov}_{p(\mathbf{Y},\mathbf{C}|\mathbf{X};\mathbf{W})}\big[\operatorname{E}_{p(\mathbf{p}_{2D}|\mathbf{Y},\mathbf{C})}[\mathbf{p}_{2D}]\big]}_{\text{Epistemic uncertainty}} \\&\approx \operatorname{Cov}\big[\{\boldsymbol{\mu}_{\mathbf{p}_{2D}}^s\}_{s=1}^S\big],    
\end{split}
 \label{modeluncertain}
\end{equation}
which is sample covariance matrix. To obtain a scalar quantification of uncertainty, we use the trace of the covariance matrix for both the aleatoric and epistemic uncertainty. 

\subsection{Generic Body Constraints}
\label{sec:GBC}

Based on body anatomy, biomechanics, and physics, we introduce hard and soft generic body constraints on the body pose and shape parameters to characterize 3D reconstruction feasibility and encourage realistic prediction.

\begin{figure}
\vspace{-.5cm}
\begin{center}
   \includegraphics[width=.95\linewidth]{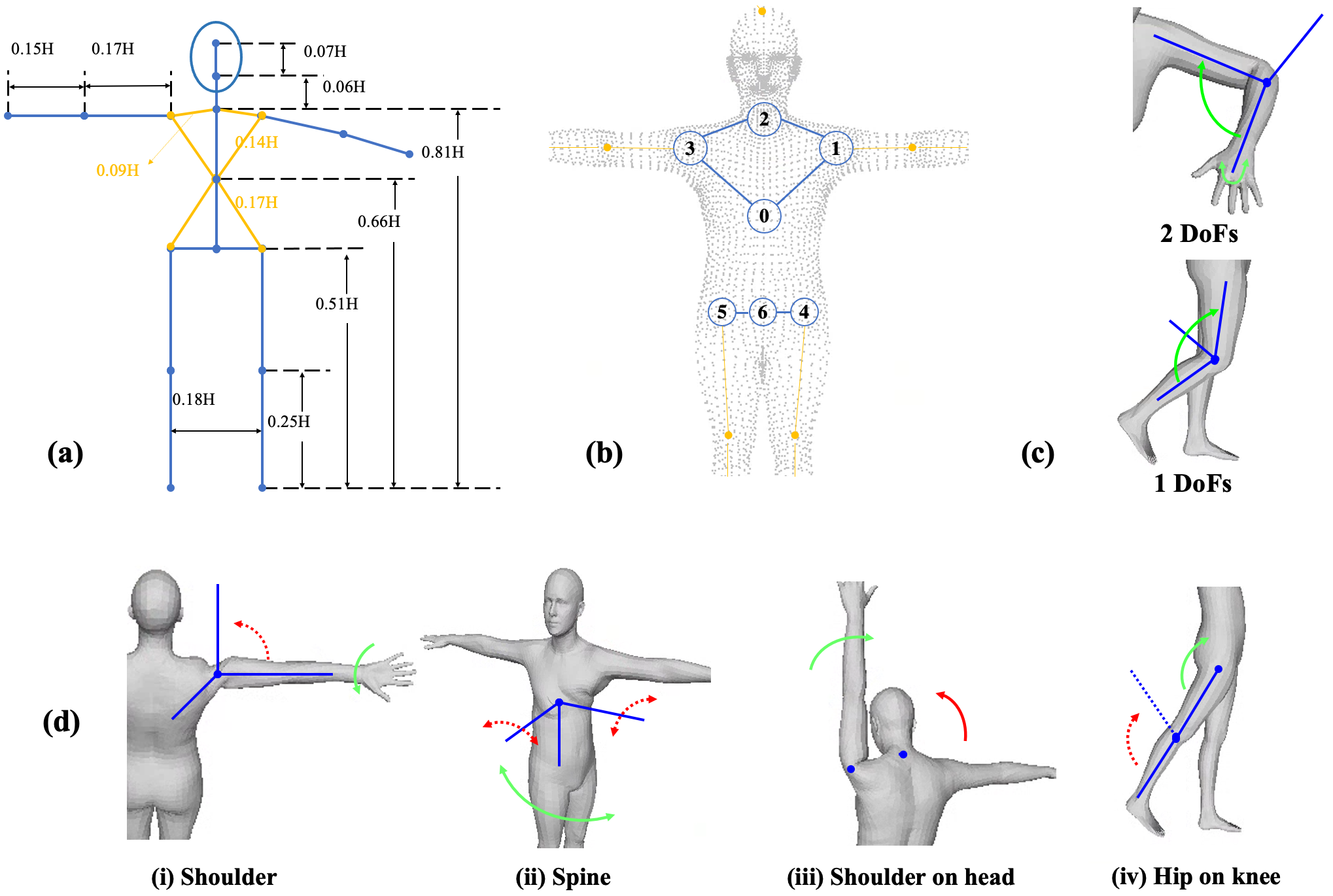}
\end{center}
\vspace{-.3cm}
   \caption{\textbf{Illustration of anatomical and  biomechanical constraints.} (a) Consistent human body proportions across individuals \cite{winter2009biomechanics}. 
   (b) Body joints that are collinear and coplanar. 
   (c) Degrees of freedom (DoFs) and movement ranges of body joints.
   (d) Examples of inter-joint dependencies. 
   }
\label{fig:generic}
\vspace{-0.28cm}
\end{figure}

\textbf{Body anatomy} studies the structure of the human body. While body symmetry and the connection of body parts by body joints are already encoded via SMPL shape bases, there are several other anatomical principles that can be exploited. Specifically, anthropometry~\cite{winter2009biomechanics} provides a set of 20 constraints on relative body proportions (Fig.~\ref{fig:generic}a), from which we formulate an anthropometric loss term:
\begin{equation}
    \mathcal{L}_{anthropometry}=\frac{1}{20}\sum_{i=1}^{20}(\frac{\hat{L}_i-L_i}{L_i})^2,
\end{equation}
where $\hat{L}$ and $L$ are the predicted and anthropometric bone lengths, respectively. Moreover, as shown in Fig.~\ref{fig:generic}b, the body torso is rigid under different actions, the shoulders, neck, and spine joints are coplanar, and the hips and pelvis joints are collinear. We find that these geometry constraints can be satisfied by imposing the anthropocentric loss. As is explicated more fully in Appx.~\ref{geom}, the geometry constraints (1) ensure realistic 3D reconstruction, especially if the body structure is not constrained; and (2) meaningfully solve the inherent depth ambiguity. Additionally, to explicitly constrain shape estimation, we add an L2 regularization term $\mathcal{L}_{beta-reg}$ on the shape parameters~\cite{loper2015smpl}, resulting in a total anatomy loss term:
\begin{equation}
    \mathcal{L}_{anatomy} = \mathcal{L}_{anthropometry} + \lambda_{beta}\mathcal{L}_{beta-reg}.
\end{equation}

\textbf{Body biomechanics}, the study of body movement mechanisms, indicates that the degrees of freedom (DoFs) and ranges of body joint are restricted \cite{nasa1995std}. For example, as illustrated in Fig.~\ref{fig:generic}c, the knee is a 1 DoF joint and the flexion can not exceed 146 degrees. The elbow is a joint with 2 DoFs that exhibits flexion and pronation/supination. The joint angle limits are naturally defined using three Euler angles, while we employ the 6D rotation representation. To impose the constraints, we recover the Euler angles from the output rotation matrix, using octant lookup based on the identified joint ranges of motion to determine the rotation order and thereby ensure a unique solution. The loss is formulated as:
\begin{equation} 
\label{eq:bioloss}
    \mathcal{L}_{biomechanics} = \sum_{i=1}^{69}( \max\{\hat{\varphi}_{i}-\varphi_{i,max},\varphi_{i,min}-\hat{\varphi}_{i}, 0\})^2,
\end{equation}
where $\{\hat{\varphi}_{i}\}_{i=1}^{69}$ are the recovered Euler angles of the 23 body joints, $\varphi_{i,max}$ and $\varphi_{i,min}$ represent the joint angle limits obtained from literature \cite{nasa1995std}, and $\max\{\cdot\}$ selects out the value that violates the bound.

From functional anatomy \cite{hamill2006biomechanical}, which studies how anatomical structure restricts joint movement, we derive additional biomechanical constraints on inter-joint dependencies. They fall into two classes: (1) those among pose parameters of an individual joint, and (2) those among pose parameters of different joints. We exemplify (1) and (2) in Fig.~\ref{fig:generic}(d,i-ii) and Fig.~\ref{fig:generic}(d,iii-iv), respectively. Specifically, (i) For shoulder joints, rotation along the arm (green arrow) limits upward movement (red arrow). (ii) For the spine joint, rotation along the torso restricts lateral and anterior-posterior rotation. (iii) Raising an arm is accompanied by an anterior translation of the head. (iv) Knee flexion is limited when the thigh is flexed. To impose the inter-joint dependency constraints, we cannot use a constant joint rotation range. Instead, we dynamically update the bounds given the current rotation and use the updated bounds to compute the loss defined in Eq.~\eqref{eq:bioloss}. For the case of Fig.~\ref{fig:generic}(d,i), denote the maximal joint angle for the shoulder joint to move up as $\varphi_{shoulder,up,max}$ and the predicted rotation angle as $\hat{\varphi}_{shoulder}$. The bound is updated according to
\begin{equation}
    \hat{\varphi}_{shoulder,up,max} = \varphi_{shoulder,up,max} - \alpha_0 \hat{\varphi}_{shoulder},
\end{equation}
where $\alpha_0$ is determined from the functional anatomy literature \cite{hamill2006biomechanical}. Other inter-joint dependencies are imposed similarly with different values of $\alpha_0$ given different underlying dependencies. The biomechanical constraints are hard constraints. Large weights are employed to encourage them to be better satisfied.

\textbf{Body physics} stipulates that different body parts can not penetrate each other. To impose the non-penetration constraints, we follow the approach introduced by \cite{Karras:2012:MPC:2383795.2383801,Tzionas:IJCV:2016}. We first detect colliding triangle pairs \cite{teschner2005collision} and measure the collision through predefined signed distance fields $\boldsymbol{\Psi}$ for each pair of colliding triangles. The loss to avoid penetration is then formulated as 
\begin{equation}
\resizebox{.7\hsize}{!}{$
\begin{split}
    \mathcal{L}_{physics} &= \sum_{(f_s,f_t)\in \mathcal{C}}
    \Big\{
    \sum_{\mathbf{v}_s\in f_s}\|-\boldsymbol{\Psi}_{f_t}(\mathbf{v}_s)\cdot\mathbf{n}_s\|^2 \\
    &+\sum_{\mathbf{v}_t\in f_t}\|-\boldsymbol{\Psi}_{f_s}(\mathbf{v}_t)\cdot\mathbf{n}_t\|^2
    \Big\},    
\end{split}
$}
\end{equation}
where $\mathcal{C}$ is the set of colliding triangles, $f_s$ and $f_t$ represent intruding and receiving triangles with the corresponding norm $\mathbf{n}_s$ and $\mathbf{n}_t$, respectively.

\textbf{The total generic loss} obtained via assembling the anatomical, biomechanical, and physics loss functions is:
\begin{equation}
\label{genericloss}
    \mathcal{L}_{generic}=\lambda_{1}\mathcal{L}_{anatomy}+\lambda_{2}\mathcal{L}_{biomechanics} +\lambda_{3}\mathcal{L}_{physics}.
\end{equation}
We highlight that these generic constraints (1) apply to \emph{all} subjects and activities, not just those represented in the training set; (2) do not require a laborious 3D data collection process; and (3) are not affected by data noise. 

\subsection{Negative Log Likelihood for 3D-2D Consistency}
\label{sec:NLL}

To further ensure that 3D prediction is consistent with the input image, we consider a novel training loss based on Negative Log Likelihood (NLL). Specifically, we formulate NLL to align 2D keypoint projections with input images $\mathbf{X}_i$ with 2D keypoint labels $\bar{\mathbf{p}}_{2D,i}\in\mathbb{R}^{J\times2}$:
\begin{align}\label{NNLkp}
    \mathcal{L}_{NLL,i}&=-\log{p(\bar{\mathbf{p}}_{2D,i}|\mathbf{X}_i;\mathbf{W})} \notag\\
    &=-\log{\iint p(\bar{\mathbf{p}}_{2D,i}|\mathbf{Y},\mathbf{C})p(\mathbf{Y},\mathbf{C}|\mathbf{X}_i;\mathbf{W})}d\mathbf{Y}d\mathbf{C} \notag\\
    &\approx -\log{\frac{1}{S}\sum_{s=1}^Sp(\bar{\mathbf{p}}_{2D,i}|\mathbf{Y}_s,\mathbf{C}_s)},
\end{align}
where $\{\mathbf{Y}_s,\mathbf{C}_s\}_{s=1}^S$ are the samples drawn from the corresponding distribution. Directly solving the integration in Eq.~\eqref{NNLkp} is intractable, so we approximate the integration through sampling. Meanwhile, we employ a reparameterization trick~\cite{kingma2013auto} to efficiently select samples in a differentiable way as
\begin{equation}
    \mathbf{Y} = \boldsymbol{\mu}_{\mathbf{Y}}(\mathbf{X};\mathbf{W}) + \mathbf{L}(\mathbf{X};\mathbf{W}) \boldsymbol{\sigma}, \ \boldsymbol{\sigma} \sim \mathcal{N}(\mathbf{0},\mathbf{I}),
    \label{repara}
\end{equation}
where $\boldsymbol{\mu}_{\mathbf{Y}}(\mathbf{X})$ is the mean and $\mathbf{I}$ is an identity matrix with the same dimension of $\operatorname{Cov}[\mathbf{Y}]$, $\mathbf{L}(\mathbf{X};\mathbf{W})$ is the Cholesky decomposition of $\operatorname{Cov}[\mathbf{Y}]$ ($\operatorname{Cov}[\mathbf{Y}]=\mathbf{L}(\mathbf{X};\mathbf{W})\mathbf{L}^T(\mathbf{X};\mathbf{W})$). In this work, we consider a diagonal covariance matrix of the body model parameters and zero covariance of the camera parameters. The mean of $p(\mathbf{p}_{2D}|\mathbf{X};\mathbf{W})$ is therefore computed using $\boldsymbol{\mu}_{\mathbf{C}}$ and the body model parameters $\mathbf{Y}=[\boldsymbol{\theta},\boldsymbol{\beta}]$ sampled using Eq.~\eqref{repara}.

\subsection{Uncertainty-guided Model Refinement Loss}
\label{sec:UncertaintyGuided}
Given that the training data span multiple datasets that vary in diversity and size (e.g.\ some contain limited data on in-the-wild scenarios with challenging poses), KNOWN augments the initial training based on the epistemic uncertainty quantified via Eq.~\eqref{modeluncertain} with a final uncertainty-guided training step. Note that the epistemic uncertainty generated by the initial training is negatively correlated with training data density, and thus explains where the model may fail due to lack of data \cite{kendall2017uncertainties}. 
We compute a refinement weight $w(\mathbf{X}_i)$ based on the quantified epistemic uncertainty $\mathcal{U}_e(\mathbf{X}_i)$ for the $i$th training data over a minibatch of size $B$:
\begin{equation}
    \label{weight}    w(\mathbf{X}_i)=1+\frac{\exp(\mathcal{U}_e(\mathbf{X}_i))}{\sum_{b=1}^B \exp(\mathcal{U}_e(\mathbf{X}_b))}.
\end{equation}
The \textbf{overall training loss} is then given by:
\begin{equation}
    \mathcal{L} =  \sum_{i=1}^N w_i\mathcal{L}_{NLL,i} +\mathcal{L}_{generic,i}, 
\end{equation}
where $w_i=\{w(\mathbf{X}_i),1\}$ with $w_i=1$ when training an initial model while $w_i=w(\mathbf{X}_i)$ during the refinement, and $N$ is the total number of training images.

\section{Experiment}
\label{experi}

We first perform ablation studies to show the effectiveness of KNOWN's body knowledge and uncertainty modeling. Then, we quantitatively compare KNOWN to the weakly-supervised state-of-the-art methods (SOTAs), \textit{i.e.} those that do not require 3D supervisions paired with input images during training. We highlight the advantages of KNOWN on the minority testing samples. Last, we qualitatively evaluate KNOWN's 3D reconstruction performance and illustrate how uncertainty modeling can lead to model improvements. This is followed by a discussion on KNOWN's generalization ability.

\noindent\textbf{Datasets and implementation.} Following the typical strategy~\cite{kanazawa2018end, kolotouros2019learning} for model training, we consider both 3D and 2D datasets. 3D datasets include Human3.6M (H36M)~\cite{ionescu2013human3} and MPI-INF-3DHP (MPI-3D)~\cite{mehta2017monocular}, which are collected in constrained environments using MoCap sysetem. 2D datasets include COCO \cite{lin2014microsoft}, LSP and LSP-Extended \cite{johnson2010clustered}, and MPII \cite{andriluka20142d}, which are diverse in poses, subjects, and backgrounds. Our implementation uses Pytorch. Body images are scaled to 224$\times$224 pixels with the aspect ratio preserved. Images from different datasets are fed into one minibatch. For Eq.~\eqref{genericloss}, we use the predicted mean of the body pose and shape parameters. For Eq.~\eqref{NNLkp}, only one sample is drawn to efficiently approximate the integration~\cite{blundell2015weight}. Appx.~\ref{implementation} provides detailed implementation settings.

\noindent\textbf{Evaluation metrics.} We evaluate the 3D
body pose estimation task using Mean Per-Joint Position Error (MPE) and MPE after rigid alignment (P-MPE) in units of millimeters; thus smaller values are better. For the evaluation on H36M, the MPE and P-MPE computations follow two typical protocols: P1 (all images) and P2 (just frontal camera images).

\subsection{Ablation Study}
\label{sec:ablation}

\begin{figure}[t]
\begin{center}
   \includegraphics[width=0.9\linewidth]{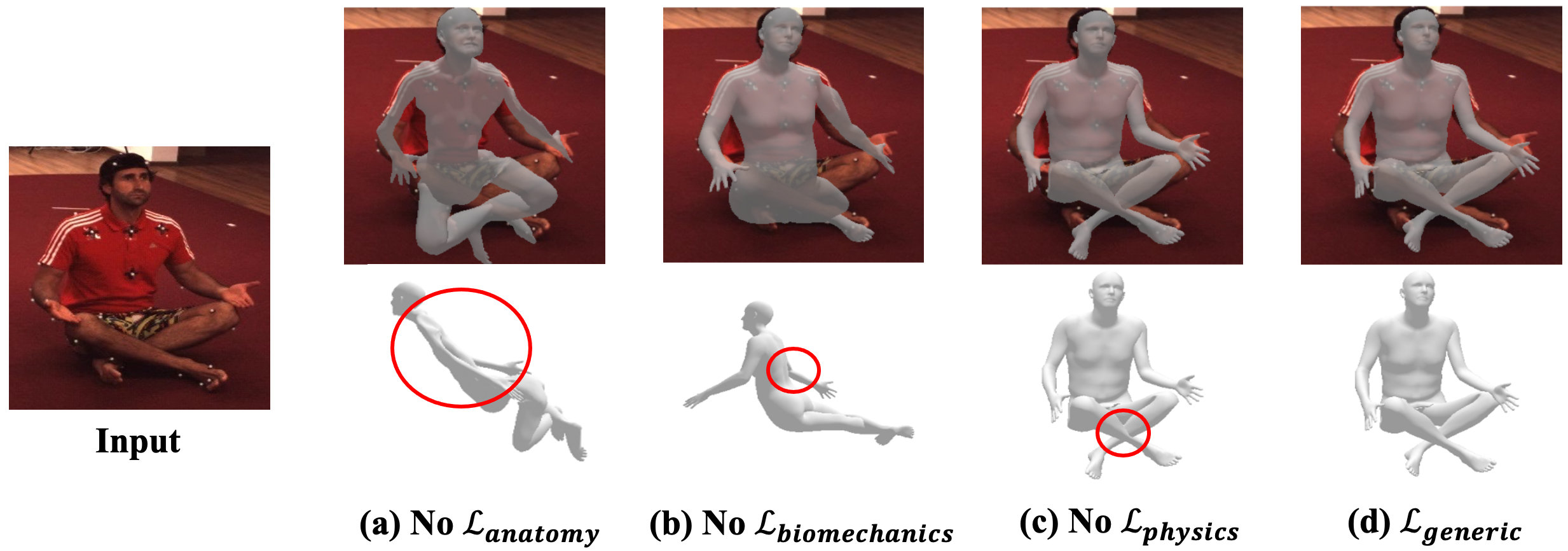}
\end{center}
\vspace{-0.4cm}
\caption{\textbf{Different generic constraints in reconstructing realistic 3D human body.} Distinct violations of the constraints are marked by red circles.}
\label{ablaqua}
\vspace{-0.2cm}
\end{figure}

\textbf{Generic constraints.} Tab.~\ref{tab:ablampjep} (Rows 1-4) and Fig.~\ref{ablaqua} summarize and illustrate the impact of the various generic constraints used by KNOWN. With all four constraints in place, the average P-MPE reconstruction error is 58.1mm. If the body anatomy constraints are removed, the predicted shape is very unreasonable on a relatively challenging image containing occlusion (Fig.~\ref{ablaqua}(a)), with large depth estimation error leading to a P-MPE of 146.0mm. Removing just the biomechanics constraint also degrades the reconstruction significantly: the P-MPE is 107.4mm and Fig.~\ref{ablaqua}(b) is a clearly invalid pose. While removing the physics constraint has a relatively small impact on P-MPE (60.7mm), it can still lead to invalid poses like that illustrated in Fig.~\ref{ablaqua}(c). Thus the three types of generic constraints are evidently synergistic, providing good, plausible 3D reconstructions from 2D body landmarks. The impact of the NLL constraint is to reduce the P-MPE somewhat, but the three generic constraints appear to be enough to ensure physical plausibility, as seen in Fig.~\ref{ablaqua}(d).

\begin{table}[t]
\tabcolsep=0.03in
\begin{center}
\begin{tabular}{ ccccc  c c }
\toprule
\multicolumn{5}{c}{Loss functions} & \multicolumn{2}{c}{P-MPE (H36M, P2)} \\
\hline
$\mathcal{L}_{ana}$ & $\mathcal{L}_{bio}$ & $\mathcal{L}_{phy}$ & $\mathcal{L}_{NLL}$ & U-refine & All & Minority \\
\midrule
 & $\checkmark$ & $\checkmark$ & $\checkmark$ & & 146.0 & 160.9  \\
$\checkmark$ & & $\checkmark$ & $\checkmark$ & & 107.4& 148.5 \\
$\checkmark$ & $\checkmark$ & & $\checkmark$ & & 60.7& 81.7   \\
\hline
$\checkmark$ & $\checkmark$ & $\checkmark$ &  & & 63.3 & 78.8 \\
\hline
$\checkmark$ & $\checkmark$ & $\checkmark$ &  $\checkmark$ &  & 58.1 &  81.3 \\
\hline
$\checkmark$ & $\checkmark$ & $\checkmark$ & $\checkmark$ & $\checkmark$ & \textbf{55.9} &  \textbf{70.3} \\
\bottomrule
\end{tabular}
\end{center}
\vspace{-0.2cm}
\caption{\textbf{Ablation study of different loss terms.} $\mathcal{L}_{ana}$, $\mathcal{L}_{bio}$, and $\mathcal{L}_{phy}$ stand for $\mathcal{L}_{anatomy}$, $\mathcal{L}_{biomechanics}$, and $\mathcal{L}_{physics}$, respectively. ``U-refine" indicates whether utilizing the uncertainty-guided refinement. When not using NLL, we use Mean Square Error (MSE) to compute the 2D keypoint reprojection error. }  
\label{tab:ablampjep}
\vspace{-0.2cm}
\end{table}

\begin{figure}
\begin{center}
   \includegraphics[width=0.99\linewidth]{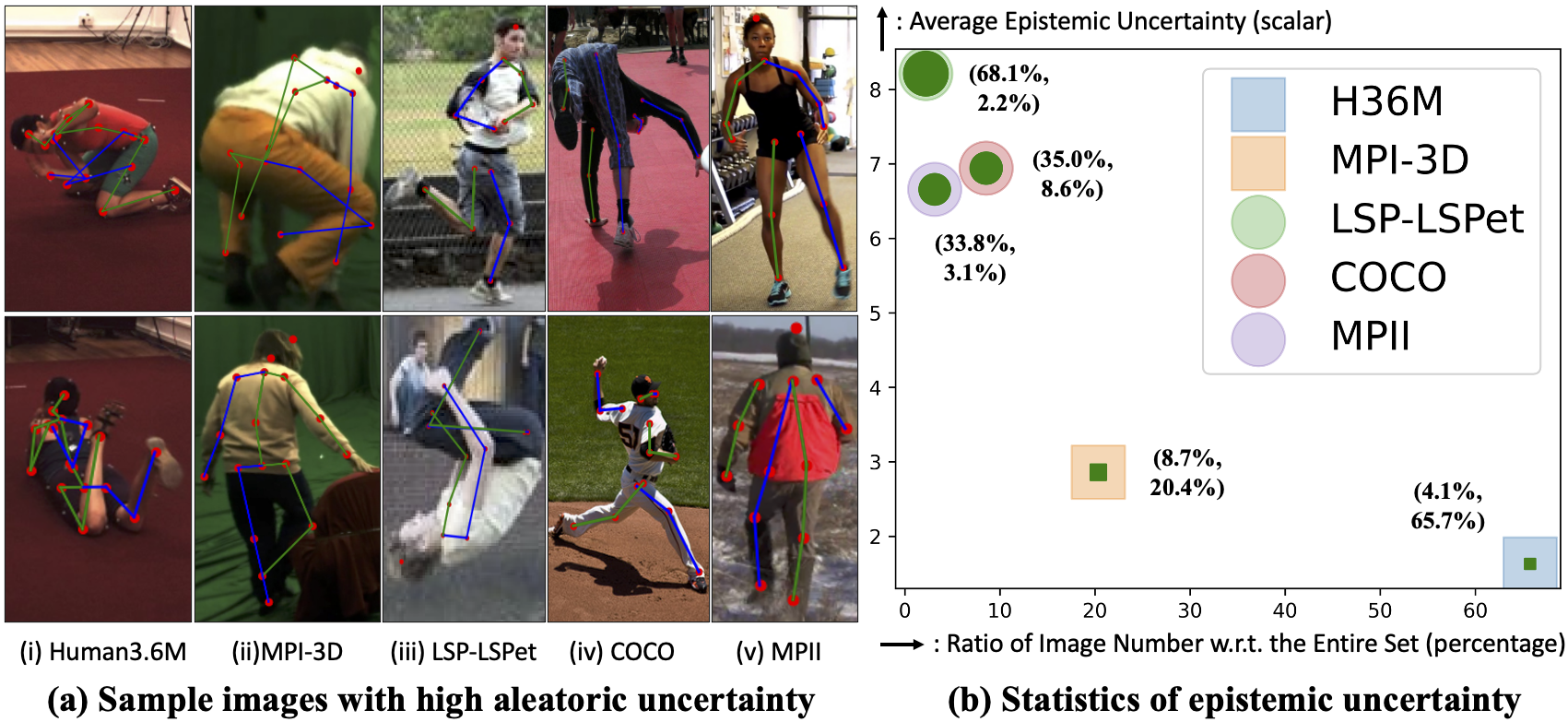}
\end{center}
\vspace{-0.4cm}
\caption{\textbf{Aleatoric and epistemic uncertainty.} (a) Aleatoric uncertainty (Eq.~\eqref{aleatoric_uncertainty}). Images with label of left (green) and right (blue) side of body skeleton are shown. (b) Epistemic uncertainty (Eq.~\eqref{epistemic_uncertainty}). The percentage of images within dataset that are minorities (top) and the percentage of images in dataset relative to entire training set (bottom) are shown alongside each dataset.}
\label{fig:uncertain}
\vspace{-0.3cm}
\end{figure}

\textbf{Uncertainty modeling.} KNOWN effectively captures both aleatoric and epistemic uncertainty. By leveraging aleatoric uncertainty, KNOWN successfully identifies the images containing large data noise due to self-occlusion (Fig.~\ref{fig:uncertain}(a), columns 1-4), poor image quality (Fig.~\ref{fig:uncertain}(a), column 3), or annotation errors (Fig.~\ref{fig:uncertain}(a), columns 2,5). For example, images in MPI-3D, collected by a markerless MoCap system, can have severe labeling errors, including inaccurate body landmark positions or mislabeling of left and right body sides (Fig.~\ref{fig:uncertain}(a),  column 2). Moreover, leveraging epistemic uncertainty, KNOWN (without refinement) characterizes data imbalance as shown in Fig.~\ref{fig:uncertain}(b). Specifically, to illustrate the data imbalance problem, we regard the images whose model uncertainty are larger than 90$\%$ of the training data as \textit{minorities}. Compared to 3D datasets (H36M, MPI-3D), 2D datasets (LSP, COCO, MPII) have smaller model size but larger minority ratio due to their greater diversity in backgrounds, poses, and subjects.

\textbf{NLL and uncertainty-guided refinement.} The NLL loss term adaptively assigns smaller weights to noisy inputs based on their larger aleatoric uncertainty (which relates to the predicted variances as derived in Eq.~(\ref{aleatoric_uncertainty})). As seen in Tab.~\ref{tab:ablampjep}, when added to the generic loss term, NLL reduces the error from 63.6mm to 58.1mm. The final step in the KNOWN pipeline is uncertainty-guided refinement, which exploits the well-captured epistemic uncertainty to refine the estimates, especially for the challenging minorities. As shown in Tab.~\ref{tab:ablampjep}, it reduces reconstruction error from 58.1mm to 55.9mm, with particular advantage on the minorities (from 81.3mm to 70.3mm).

\begin{table}[t]
\begin{center}
\tabcolsep=0.03in
\begin{tabular}{l c  c c  c }
\toprule
\multirow{2}{*}{Method} & \multirow{2}{*}{\shortstack{Source of\\the Prior}} & \multicolumn{2}{c}{H36M} & MPI-3D \\
\cline{3-5}
& & MPE & P-MPE & P-MPE \\
\midrule
HMR~\cite{kanazawa2018end} & 3D & 106.8 & 66.5 & 113.2 \\
SPIN~\cite{kolotouros2019learning}& 3D & - & 62.0 & 80.4  \\
Song \textit{et al.}~\cite{song2020human} & 3D & - & 56.4 & - \\
Kundu \textit{et al.}~\cite{kundu2020appearance} & 3D & 86.4 & 58.2 & - \\
HUND~\cite{zanfir2021neural} & 3D & 91.8 & - & -\\
THUNDER~\cite{zanfir2021thundr} & 3D & 87.0 & 59.7 & - \\
PoseNet~\cite{tripathi2020posenet3d} & 2D & - & 59.4 & 102.4 \\
Yu \textit{et al.}~\cite{yu2021skeleton2mesh} & 2D & 87.1 & - & 87.4  \\
\midrule
Ours & Knowledge & \textbf{79.2} & \textbf{55.9} & \textbf{79.3}
 \\
\bottomrule
\end{tabular}
\end{center}
\caption{\textbf{Comparisons to SOTAs.} Existing works rely on data-drive priors learned from 3D MoCap data (3D) or 2D body pose data (2D), while we derive generic prior from body knowledge (Knowledge). Results of other works are from their papers.}
\label{cross}
\end{table}

\begin{table}[t]
\begin{center}
\tabcolsep=0.06in
\begin{tabular}{ l c  c }
\toprule
Method & All & Minority \\
\midrule
HMR~\cite{kanazawa2018end} (w.o./w. $\mathcal{P}$) & 65.2/57.2 & 108.5/107.3 \\
SPIN~\cite{kolotouros2019learning} (w.o./w. $\mathcal{P}$) & 62.0/41.1 & -/71.0 \\
\midrule
Ours & \textbf{55.9} &  \textbf{70.3} \\
\bottomrule
\end{tabular}
\end{center}
\caption{\textbf{Qualitative evaluation on the minorities.} The numbers are P-MPE evaluated on the testing images of H36M (P2). $\mathcal{P}$ denotes the usage of the 3D annotations paired with the input images.
} 
\label{tab:minority}
\vspace{-0.28cm}
\end{table}

\begin{figure}[t]
\begin{center}
  \includegraphics[width=.99\linewidth]{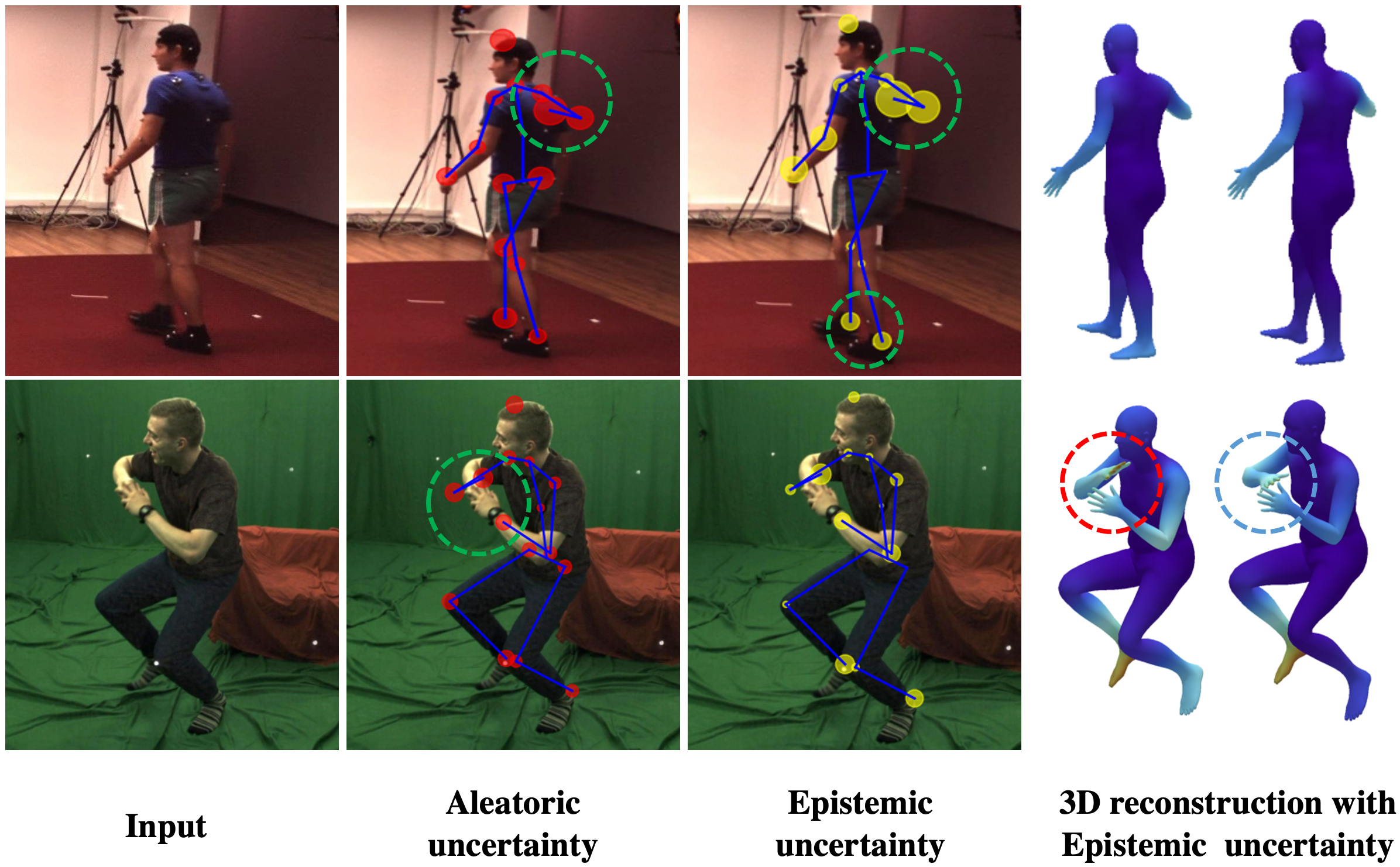}
\end{center}
\vspace{-0.3cm}
  \caption{\textbf{Qualitative evaluation with uncertainty visualization.} Images are the testing set from: H36M (top row), and MPI-3D (bottom row). In columns 2-3, the size of the red (aleatoric) and yellow (epistemic) circles indicate the level of quantified uncertainty. For the 3D reconstruction results before (columns 4) and after the refinement (columns 5), the color of each vertex represents the level of epistemic uncertainty on each vertex with lighter colors representing higher epistemic uncertainty.}
\label{quali}
\vspace{-0.2cm}
\end{figure}

\subsection{Comparison to SOTAs}

Tab.~\ref{cross} compares KNOWN with prior works. In order to ensure a fair comparison, we exclude methods that rely on additional 2D features as input~\cite{sengupta2020synthetic,sengupta2021probabilistic,sengupta2021hierarchical,gong2022self}. KNOWN achieves the best performance on both H36M and MPI-3D. In particular, HMR adopts a data-driven prior learned from 3D MoCap data (including H36M but not MPI-3D) as a regularization for training with 2D body landmarks only. HMR fails to perform well on MPI-3D, where the poses are different from the learned prior. In contrast, KNOWN leverages the generic constraints to achieve consistent performance on both of the datasets. Building upon HMR, SPIN further incorporates an optimization procedure into training to iteratively refine the estimation. Instead of treating all the samples equally and refining the model via multiple rounds as SPIN, KNOWN effectually targets the challenging minority images by employing its uncertainty-guided refinement strategy, outperforming SPIN substantially on H36M and slightly on MPI-3D. Moreover, Kundu \textit{et al.} depend on a 3D pose prior and additional self-supervision from appearance cues for training. HUND and THUNDER build prior models from MoCap data to constrain the reconstruction space. Yu \textit{et al.} and PoseNet extract a 2D projection prior from 2D pose data to encourage prediction feasibility, while Yu \textit{et al.} relies on an extra inverse kinematic mapping module for incorporating body knowledge. Compared to them, KNOWN achieves better performance by (1) leveraging the generic body constraints, which are systematic, physically meaningful, and readily incorporated into different frameworks, and (2) using uncertainty modeling, thereby achieving more efficient and effective training.

To further demonstrate the advantages of KNOWN, we highlight the evaluation on the minorities. Tab.~\ref{tab:minority} reports the evaluation of KNOWN on the minority testing images of H36M compared to two recent works. For both HMR and SPIN, performance drops severely on the minorities even with the usage of 3D annotations (paired). In contrast, KNOWN outperforms their fully supervised setting without requiring any 3D information. For example, we achieve 70.3mm, significantly better than HMR's 107.3mm. The evaluation on minorities further establishes the importance of uncertainty modeling and generic knowledge and demonstrates the advantages of KNOWN.

\subsection{Qualitative Evaluation}

\textbf{3D reconstruction with uncertainty visualization.} As shown in Fig.~\ref{quali}, KNOWN generates accurate 3D reconstruction results. Beyond 3D reconstruction, KNOWN also provides both aleatoric and epistemic uncertainty of the prediction. The estimated uncertainty captures the hard regions, such as large aleatoric and epistemic uncertainty at the occluded regions or at the leaf joints (regions circled by green in Fig.~\ref{quali}). By further incorporating uncertainty-guided refinement, KNOWN improves accuracy for these challenging and low confidence cases (region circled by blue over region circled by red in Fig.~\ref{quali}). Moreover, besides estimating the uncertainty in 2D keypoint projections, KNOWN captures the epistemic uncertainty of each 3D vertex prediction (columns 4-5 of Fig.~\ref{quali}), providing different insights into the model's behaviour as discussed in Appx.~\ref{compute_uncertainty}.

\begin{figure}[t]
\begin{center}
   \includegraphics[width=1\linewidth]{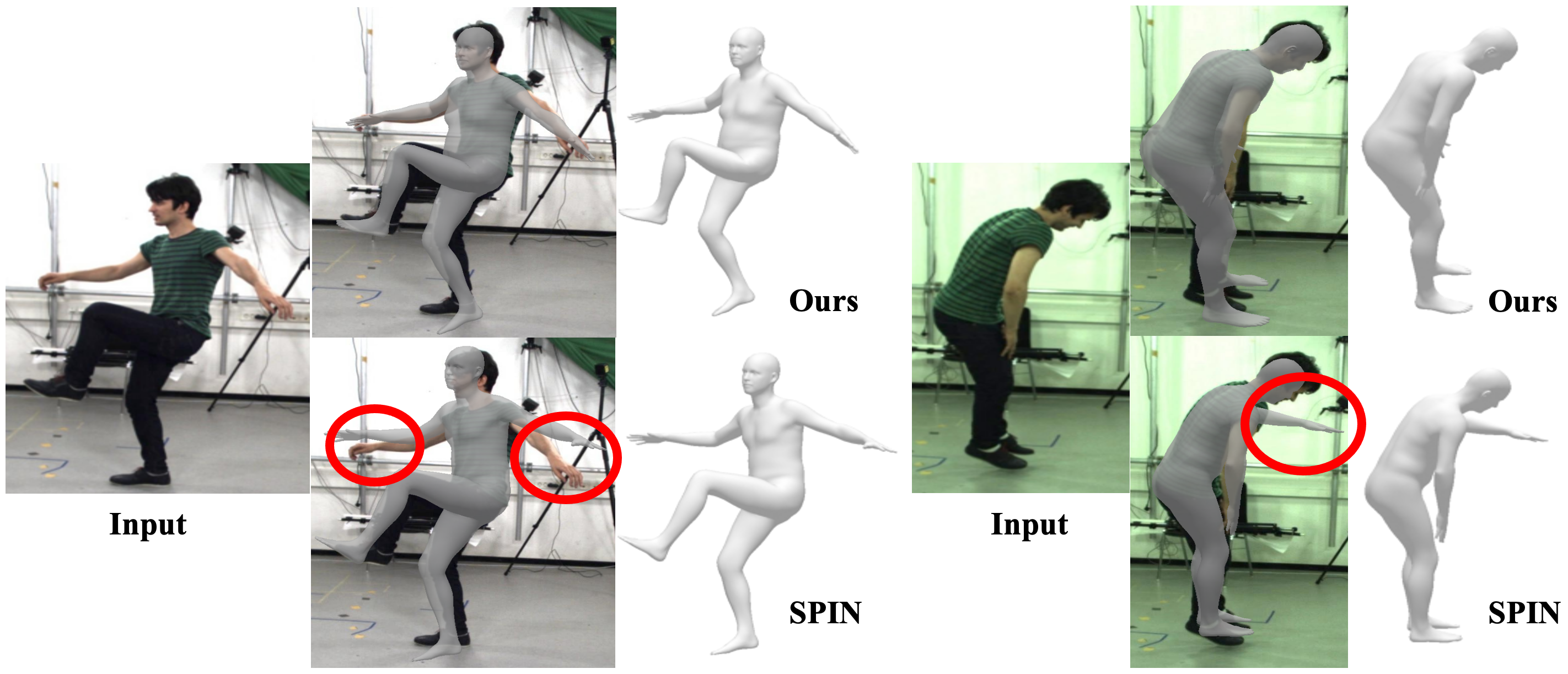}
\end{center}
\vspace{-0.3cm}
\caption{\textbf{Qualitative evaluation with comparison of using the generic prior (KNOWN) and the data-driven prior (SPIN).}}
\label{ablaspin}
\vspace{-0.2cm}
\end{figure}

\textbf{Generalization.} Fig.~\ref{ablaspin} compares KNOWN with the data-driven SPIN method on testing images from MPI-3D. As illustrated in these examples, SPIN can fail by generating results that, while plausible in 3D pose, are not well-aligned with the input images (regions circled by red in Fig.~\ref{ablaspin}). In contrast, KNOWN generalizes better to different data sets because it utilizes generic body knowledge from literature rather than data.

\section{Discussion}

Thus far we have focused primarily on the benefits of KNOWN considering the challenging scenario where 3D annotations are unavailable. As described more fully in Appx.~\ref{differentanno}), KNOWN can also utilize paired 3D annotations. Here we compare KNOWN with prior works that use uncertainty modeling in conjunction with 3D annotations.

Tab.~\ref{paired} shows that KNOWN compares favorably to recent methods on H36M and 3DPW~\cite{vonMarcard2018} (a dataset captured outdoors with more diverse backgrounds). Compared to the deterministic data-driven baselines (HMR and SPIN) and SOTAs with uncertainty modeling (Biggs \textit{et al.} and ProHMR), KNOWN achieves better performance on both datasets without using additional MoCap data. Moreover, existing uncertainty modeling approaches only capture aleatoric uncertainty, which is used solely during testing to generate plausible hypotheses. In contrast, KNOWN captures both aleatoric and epistemic uncertainty and improves the training process by utilizing the well-captured uncertainty. Tab.~\ref{paired} illustrates that KNOWN achieves these advantages while sacrificing very little in terms of memory footprint and speed, consuming just 103.9MB in model size with a near-real-time running speed of 11.6ms. This suggests that it would be quite practical in a variety of real-world applications, such as in safety-critical scenarios that also require
accurate
uncertainty quantification.

\begin{table}
\begin{center}
\tabcolsep=0.04in
\begin{tabular}{l c c c cc }
\toprule
\multirow{2}{*}{Methods} &   \multirow{2}{*}{$\mathcal{U}$}  & \multirow{2}{*}{$\mathcal{M}$} & \multirow{2}{*}{$\mathcal{S}$} & \multicolumn{2}{c}{P-MPE} \\
\cline{5-6}
& & &  & H36M & 3DPW \\
\hline
HMR \cite{kanazawa2018end} (w. $\mathcal{P}$) & N/A & 103.1 & 10.2 & 56.8 & 81.3 \\
SPIN \cite{kolotouros2019learning} (w. $\mathcal{P}$) & N/A & 103.1 & 10.2 & 41.1 & 59.1  \\
\hline
Bigss \textit{et al.}\cite{biggs20203d} & $\mathcal{U}_a$ & 566.6 & - & 41.6 & 59.9 \\
ProHMR \cite{kolotouros2021probabilistic} & $\mathcal{U}_a$ & 223.0 & 16.4 & 41.2 & 59.8 \\
\hline
Ours  & \multirow{2}{*}{\underline{$\mathcal{U}_a$+$\mathcal{U}_e$}} & \multirow{2}{*}{103.9} & \multirow{2}{*}{11.6} & 55.9 & 70.4  \\
Ours (w. $\mathcal{P}$) & &  & & \textbf{40.5} & \textbf{58.1}  \\
\bottomrule
\end{tabular}
\end{center}
\caption{\textbf{Evaluation of using paired 3D annotations.} $\mathcal{U}$ indicates the ability to capture aleatoric ($\mathcal{U}_a$) or epistemic uncertainty ($\mathcal{U}_e$). The unit of model memory ($\mathcal{M}$) and running speed ($\mathcal{S}$) is MB and ms, respectively. The P-MPE computation on H36M follows P2. Note that KNOWN not only captures uncertainty, but leverages it effectively to improve the model.}
\label{paired}
\vspace{-0.28cm}
\end{table}

\section{Conclusion}

We have introduced KNOWN, a knowledge-encoded probabilistic model that tackles the data insufficiency problem in 3D body reconstruction. KNOWN introduces a set of constraints derived from a systematic study of body knowledge available from literature. These constraints are generalizable, explainable, and easy to adapt to different frameworks. Moreover, KNOWN is a new probabilistic framework that efficiently and effectively captures both aleatoric and epistemic uncertainty. The captured uncertainty handles the data noise and data imbalance through training with a robust NLL loss and a novel uncertainty-guided refinement strategy. KNOWN achieves remarkable performance without relying on any 3D data, and is efficient in memory footprint and computation speed, suggesting that it would be useful in a wide variety of applications --- particularly those for which acquiring 3D data is virtually impossible, such as 3D animal body pose reconstruction.

\paragraph{Acknowledgement} This work is supported in part by the Cognitive Immersive Systems Laboratory (CISL), a collaboration between IBM and RPI, and also a center in IBM's AI Horizon Network.

\clearpage
{\small
\bibliographystyle{ieee_fullname}
\bibliography{main.bbl}
}

\clearpage
\appendix

In this Appendix, we first introduce additional details referred in the main manuscript which include
\begin{itemize}
    \item Section~\ref{geom}: the way of encoding the geometry constraints and its significance;
    \item Section~\ref{implementation}: additional introduction of the datasets and implementation;
    \item Section~\ref{compute_uncertainty}: the method of quantifying the uncertainty of 3D mesh vertex prediction;
    \item Section~\ref{differentanno}: implementation details of incorporating additional annotations.
\end{itemize}
Then we provide additional experiment results including
\begin{itemize}
    \item Section~\ref{appen:minoritytesting}: example minority images in training and testing sets;
    \item Section~\ref{AddiQualitative}: additional qualitative evaluation;
    \item Section~\ref{shapeestimation}: evaluation results of shape estimation;
    \item Section~\ref{labelingnoise}: measuring the labeling noise presented in existing MoCap data through the proposed generic constraints.
\end{itemize}

\section{Geometry Constraints}
\label{geom}

As illustrated in Figure~2b of the main manuscript, the shoulders, neck, and spine joints are coplanar; the hips and pelvis joints are collinear. Let $\mathbf{P}_{i}$ be the 3D position of joint $i$ and $\mathbf{P}_{ij}=\mathbf{P}_{j}-\mathbf{P}_{i}$ be the bone vector, where $i,j=0,...,6$ and $i\neq j$. The geometry constraints can be imposed via encouraging the angle between bone $\mathbf{P}_{64}$ and $\mathbf{P}_{65}$ to be 180 degrees, and the angle between bone $\mathbf{P}_{02}$ and the norm of plane $\mathbf{P}_{0,1,3}$ to be 90 degrees. The losses can be formulated accordingly as 
\begin{align}
    \mathcal{L}_{coplanar} &= \frac{|(\mathbf{P}_{01}\times\mathbf{P}_{03})\cdot\mathbf{P}_{02}|}{\|\mathbf{P}_{01}\times\mathbf{P}_{03}\|\|\mathbf{P}_{02}\|},
    \\
    \mathcal{L}_{collinear} &=\frac{|\mathbf{P}_{64}\times\mathbf{P}_{65}|}{\|\mathbf{P}_{64}\|\|\mathbf{P}_{65}\|}, \\
    \mathcal{L}_{geometry} &=  \mathcal{L}_{coplanar} + \lambda_{colinear}
    \mathcal{L}_{collinear}.
\end{align} 

The geometry constraints are derived based on the knowledge of human body structure. In Table~\ref{tab:abla} row 2, we demonstrate that imposing the geometry constraints ensures realistic upper body reconstruction (the reconstructed joints becomes colinear and coplanar). Moreover, these geometry characteristics can be exploited to solve the inherent depth ambiguity in lifting 2D observation to its 3D configuration. Specifically, let $\mathbf{P}_{1}$, $\mathbf{P}_{2}$, and $\mathbf{P}_{3}$ be the 3D position of three collinear points in the camera coordinate system. Under perspective projection, we have 
\begin{equation}
\label{perspective}
    \lambda \mathbf{p}_{i} = \mathbf{K} \mathbf{P}_{i}, 
\end{equation}
where $i=1,2,3$, $\mathbf{K}$ is the camera intrinsic matrix, and $\lambda$ is a scalar. For Equation~\ref{perspective}, $\mathbf{P}_{i}$ can not be uniquely solved due to the depth ambiguity (the 2D-3D correspondences provide two equations but with three unknowns). However, when the three points are collinear, two additional equations can be introduced through
\begin{equation}
    \frac{\mathbf{P}_{1}-\mathbf{P}_{2}}{\|\mathbf{P}_{1}-\mathbf{P}_{2}\|} = \frac{\mathbf{P}_{1}-\mathbf{P}_{3}}{\|\mathbf{P}_{1}-\mathbf{P}_{3}\|}.
\end{equation}
Given the camera intrinsic parameters and the 3D distance between the three points, unique values of $\mathbf{P}_{1}$, $\mathbf{P}_{2}$, and $\mathbf{P}_{3}$ can be solved. Similarly, the coplanarity can introduce additional constraints for alleviating the depth ambiguity. As we do not assume the camera information is given, further study of the geometry constraints is not discussed here.

\begin{table*}
\begin{center}
\vspace{-0.2cm}
\begin{tabular}{l | c  c  c  c  c | c  c }
\toprule
\multirow{3}{*}{\textbf{Models}} & \multicolumn{5}{c|}{Constraint Satisfaction} & \multicolumn{2}{c}{Reconstruction Error} 
\\ 
\cline{2-8}
& \multicolumn{2}{c}{Anatomy}
& \multicolumn{2}{c}{Biomechanics} & \multirow{2}{*}{Physics} & \multirow{2}{*}{MPE} & \multirow{2}{*}{P-MPE} \\
& Bone & Geometry & Angle & Angle-inter & \\
\hline
$\mathcal{L}_{NLL} $ 
& 101.2 & 83.2/166.5 & 30.7/30.8 & 109.1 & 97 & 296.5 & 161.2 \\ 
 $\mathcal{L}_{NLL}+\mathcal{L}_{geometry}$ &
130.8 & 89.4/178.7 & 41.6/41.7 & 35.2 & 100 & 415.3 & 311.5\\ 
\hline
3DPW GT 
 & 21.0
 & 89.8/178.6 
 & 4.1/4.1 & 6.6
 & 2  & - & - \\
\bottomrule
\end{tabular}
\end{center}
\caption{\textbf{Evaluation of the constraints satisfaction on the ground truth data and  the usage of the geometry constraints.} For quantifying the constraint satisfaction, we compute the mean per bone length error (Bone, in mm), average angle induced by the coplanar/colinear joints (Geometry, in degrees), mean per joint angle violations with/without considering the inter-joint dependency (Angle/Angle-inter, in degrees), and percentage of data with penetration (Physics, in percents). The bone and geometry constraints are soft constraints, while the biomechanic and physic constraints are hard constraints that should be strictly satisfied. The angle induced by the coplanar and coplinear joints should approximate to 90 and 180 degrees, respectively.}
\label{tab:abla}
\vspace{-0.3cm}
\end{table*}

\section{Datasets and Implementation Details}
\label{implementation}

\noindent\textbf{Datasets.} H36M includes 5 subjects performing 15 daily actions like Eating, Greeting, and \textit{et al}, and it consists of a total of 312,188 training images. MPI-3D includes 8 subjects covering 8 typical action classes like Exercise, Sitting, and \textit{et al.}, with a total of 96,620 valid training images. COCO is a dataset widely used for segmentation and detection tasks. LSP and LSP-Extended (10,482 images) and MPII (14,806 images) are standard datasets for 2D pose estimation and involve more diverse poses than COCO. 

\noindent\textbf{Implementation.} ResNet-50 \cite{he2016identity} is pretrained on the ImageNet classification. The images from different datasets are fed into one minibatch with the following split: H36M (0.35), MPI-INF-3DHP (0.1), COCO (0.35), MPII (0.1), and LSP and LSP-Extended (0.1). The input images are augmented with random scaling and flipping. The model is trained using Pytorch's Adam solver with a learning rate of $10^{-5}$ and weight decay $10^{-4}$. The training is conducted on one 2080Ti GPU with batch size of 64. During training, we observe that it is efficient to train the regression model by first encoding the generic prior. We hence first train the regression model on H36M~\cite{ionescu2013human3} for $\sim150K$ iterations and then continue training on all the datasets for $\sim500K$ iterations. Once the initial model is trained, we quantify the uncertainty of all the training samples and compute the corresponding uncertainty-guided refinement weights. Based on the computed weights, we further refine the initial model for $\sim100K$ iterations and obtain the final model. 

Regarding the hyperparameters, we use 10 for the keypoints reprojection loss, 500 for the body anatomy loss, 1000 for the the biomechanics loss, 1000 for physic loss, and 1 for scaling the refinement weights. We also add regularization on the trace of the predicted covariance matrix of the 2D keypoint projection with a weight of 50. This term encourages the model to converge to the position with small 2D projection error.

When calculating the model memory, we use Pytorch's \texttt{model.parameters()} and \texttt{model.buffers()} to count all the parameters and buffers stored in a model. When evaluating the model speed, we perform the model inference on a computer with a Intel(R) Xeon(R) W-2135 CPU and one 2080Ti GPU.

\section{Uncertainty Quantification for 3D Mesh Vertex Prediction}
\label{compute_uncertainty}

Not limited to quantifying the uncertainty of 2D body keypoint prediction, KNOWN can also quantify the epistemic uncertainty of the 3D vertex prediction. Specifically, for 3D body mesh vertices $\mathbf{M}$, its conditional probability given 3D body model parameters $\mathbf{Y}$ follows Gaussian distribution: 
\begin{align}
    p(\mathbf{M}|\mathbf{Y}) &= \mathcal{N}\big(\boldsymbol{\mu}_{\mathbf{M}}(\mathbf{Y}),\boldsymbol{\Sigma}_{\mathbf{M}}(\mathbf{Y})\big), \label{ditribution3d1}
\end{align}
where the mean of the Gaussian distributions are specified by the body pose and shape parameters via the forward kinematic process. We quantify the epistemic uncertainty of the 3D vertex prediction as 
\begin{equation}
    \begin{split}
   \underbrace{\operatorname{Cov}_{p(\mathbf{Y}|\mathbf{X};\mathbf{W})}\big[\operatorname{E}_{p(\mathbf{M}|\mathbf{Y})}[\mathbf{M}]\big]}_{\text{Epistemic uncertainty}} &= \operatorname{Cov}_{p(\mathbf{Y}|\mathbf{X};\mathbf{W})}\big[ \boldsymbol{\mu}_{\mathbf{M}}(\mathbf{Y})\big].
    \end{split}
    \label{eq:3Duncertainty}
\end{equation}
Directly computing the right side of Equation~\ref{eq:3Duncertainty} is difficult. We approximate the value via sample covariance: 
\begin{equation}
 \underbrace{\operatorname{Cov}_{p(\mathbf{Y}|\mathbf{X};\mathbf{W})}\big[\operatorname{E}_{p(\mathbf{M}|\mathbf{Y})}[\mathbf{M}]\big]}_{\text{Epistemic uncertainty}} \approx
 \operatorname{Cov}\big[\{\boldsymbol{\mu}_{\mathbf{M}}^s\}_{s=1}^S\big],
\end{equation}
where $\{\boldsymbol{\mu}_{\mathbf{M}}^s\}_{s=1}^S$ is computed using the samples of $\mathbf{Y}$. 

For the visualization of the 3D vertex prediction uncertainty in Figure~5 of the main manuscript, the vertex color represents the epistemic uncertainty computed following the method introduced above. The colors are obtained from a standard color map after normalizing the scalar uncertainty value of each vertex to a range from 0 to 1.

Compared to the uncertainty quantified on the 2D keypoint projections, the epistemic uncertainty quantified on 3D vertex prediction does not involve the projection process. Future work can consider further distinguish these two types of uncertainty to account for the uncertainty in estimating the camera parameters.

\begin{figure}[t]
\begin{center}
   \includegraphics[width=0.95\linewidth]{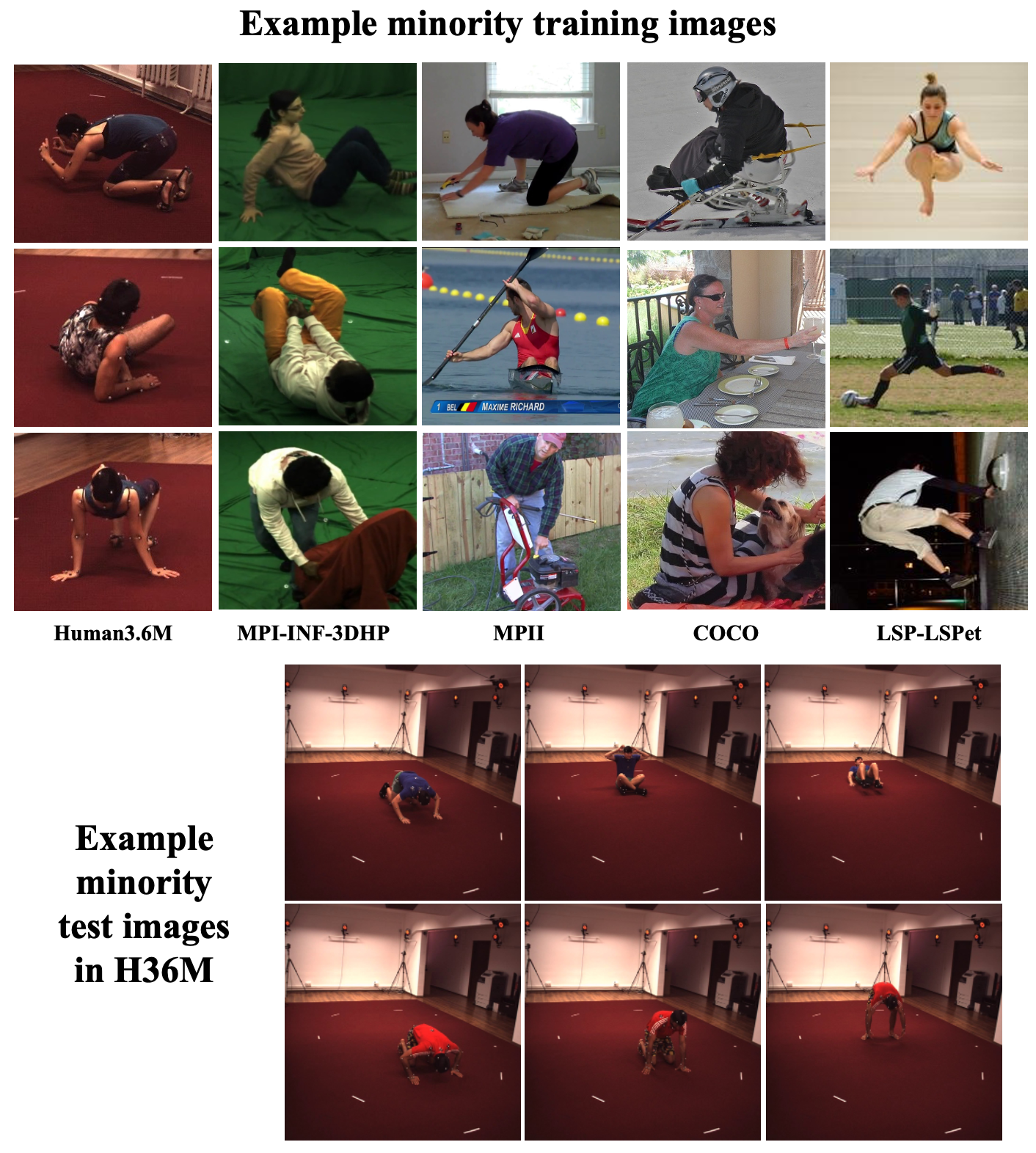}
\end{center}
\vspace{-0.2cm}
\caption{\textbf{Example minority images.}}
\label{fig:minoritytesting}
\vspace{-0.2cm}
\end{figure}

\begin{figure*}[t]
\begin{center}
   \includegraphics[width=0.99\linewidth]{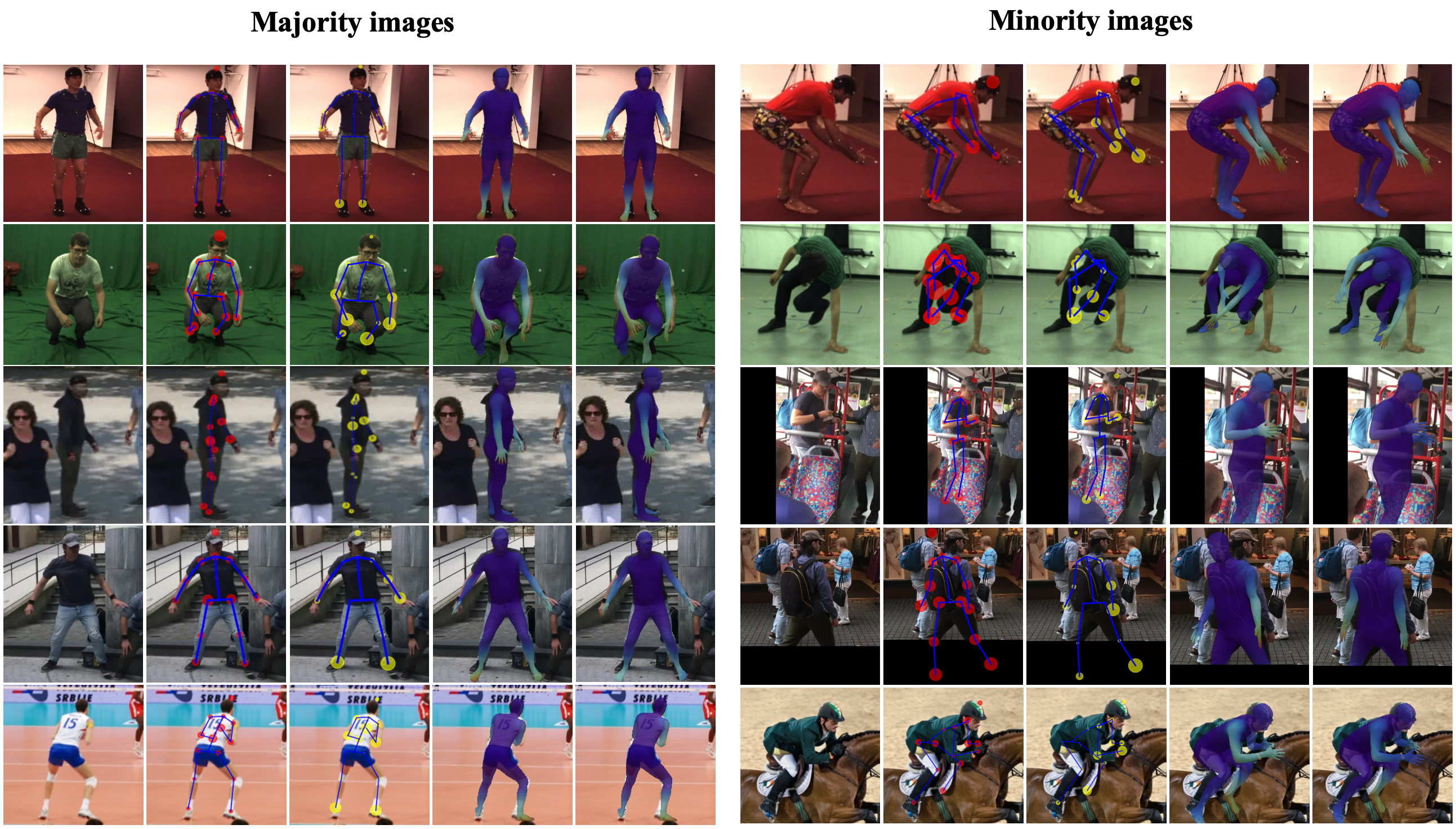}
\end{center}
   \caption{\textbf{Qualitative evaluation on the majorities and minorities.} The test images from top to bottom are from H36M (row 1), MPI-3D (row 2), 3DPW (row 3-4), and LSP (row 5), respectively. The images from lift to right are the input image, data uncertainty (without refinement), model uncertainty (without refinement), and 3D reconstruction results (without/with refinement), respectively.}
\label{supresult}
\end{figure*}

\section{Utilizing Additional Annotations}
\label{differentanno}

Training of KNOWN can easily incorporate annotations from different sources when they are available. Specifically, additional annotations can be incorporated during model finetuning via minimizing a corresponding reconstruction error. We here discuss the usage of the paired 3D annotations.

Paired 3D annotations indicate either 3D body joint position annotation or 3D body model parameter annotation that are paired with an image. Paired 3D annotations are hard to obtain and they are typically collected indoors. For the employed training datasets, H36M, MPI-3D, COCO, MPII, and LSP and LSP-Extended, only H36M and MPI-3D include the 3D body joint position annotations. To incorporate these annotations, we formulate the following loss functions:
\begin{equation}
     \mathcal{L}_{pair3D}=\|\mathbf{P}-\mathbf{\hat{P}(\boldsymbol{\mu}_{\theta},\boldsymbol{\mu}_{\beta})}\|_2^2, 
\end{equation}
where $\mathbf{\hat{P}(\boldsymbol{\mu}_{\theta},\boldsymbol{\mu}_{\beta})}$ is the predicted 3D body joint position computed based on the mean pose and shape estimates. The overall training loss is then calculated as:
\begin{equation}
    \mathcal{L} = \sum_{i=1}^N w_i(\mathcal{L}_{NLL,i}+\mathcal{L}_{pair3D}) + \mathcal{L}_{generic,i}.
\end{equation}
The overall training loss includes the uncertainty-guided refinement weights to effectively leverage the data from a specific domain based on their uncertainty.

\section{Example Minority Images}
\label{appen:minoritytesting}

In Figure~\ref{fig:minoritytesting}, we present example minority images from different training datasets and the testing set of H36M (Protocol 2). The minorities are mainly the images with large camera angle, severe occlusion, or extreme poses --- situations that make
their reconstruction particularly challenging. KNOWN successfuly improve model performance on these challenging image via uncertainty-guided refinement.

\section{Additional Qualitative Evaluation}
\label{AddiQualitative}

In Figure~\ref{supresult}, we present additional qualitative evaluation on the majorities (small epistemic uncertainty) and minorities (large epistemic uncertainty). The majorities in each test set are the images with large data density. As shown, the majorities typically posses simple poses and few occlusion. The model performance with and without employing the refinement are similar on the majorities. By contrast, the minorities in each test set are the images with low data density and they always contain more challenging poses and severe occlusion. Applying the uncertainty-guided refinement loss shows significant improvements on the minorities. Specifically, the figures at the last row of Figure~\ref{supresult} are the evaluation on a minority image from LSP's test set. The left leg and the two arms of the person in the input image are occluded or blurred with the backgrounds. As a result, our model shows large epistemic uncertainty on these regions. Moreover, utilizing the uncertainty-guided refinement improves the model performance on these challenging cases, such as the left arm aligns better with the image.

\section{Additional Quantitative Evaluation on Body Shape Estimation}
\label{shapeestimation}
We demonstrate KNOWN's improved body shape estimation performance via six-part body segmentation accuracy evaluated on the LSP test set, following the typical evaluation protocol used by \cite{kanazawa2018end}. During evaluation, body part segmetations are obtained by rendering the 3D prediction on image using the predicted camera parameters. Without using any 3D information, KNOWN's body part segmentation retains accuracy of 87.40 and F1 score of 0.65, which are better than HMR's 87.40 and 0.59, respectively.

\begin{table}
\begin{center}
\begin{tabular}{l | c c}
\toprule
\multirow{2}{*}{\textbf{LSP}}  & \multicolumn{2}{c}{Parts}  \\
\cline{2-3} & Acc. & F1 \\
\hline
HMR \cite{kanazawa2018end} & 87.00 & 0.59\\
\hline
Ours & \textbf{87.40} & \textbf{0.65}  \\
\bottomrule
\end{tabular}
\end{center}
\caption{\textbf{Evaluation of body shape estimation.} The accuracy (Acc.) and F1 score (F1) are computed on six-part body segmentation on LSP's test set.}
\label{LSP}
\end{table}

\section{Measuring Labeling Noise Presented in Existing MoCap Datasets}
\label{labelingnoise}

\begin{figure}[t]
    \centering
    \includegraphics[width=0.95\linewidth]{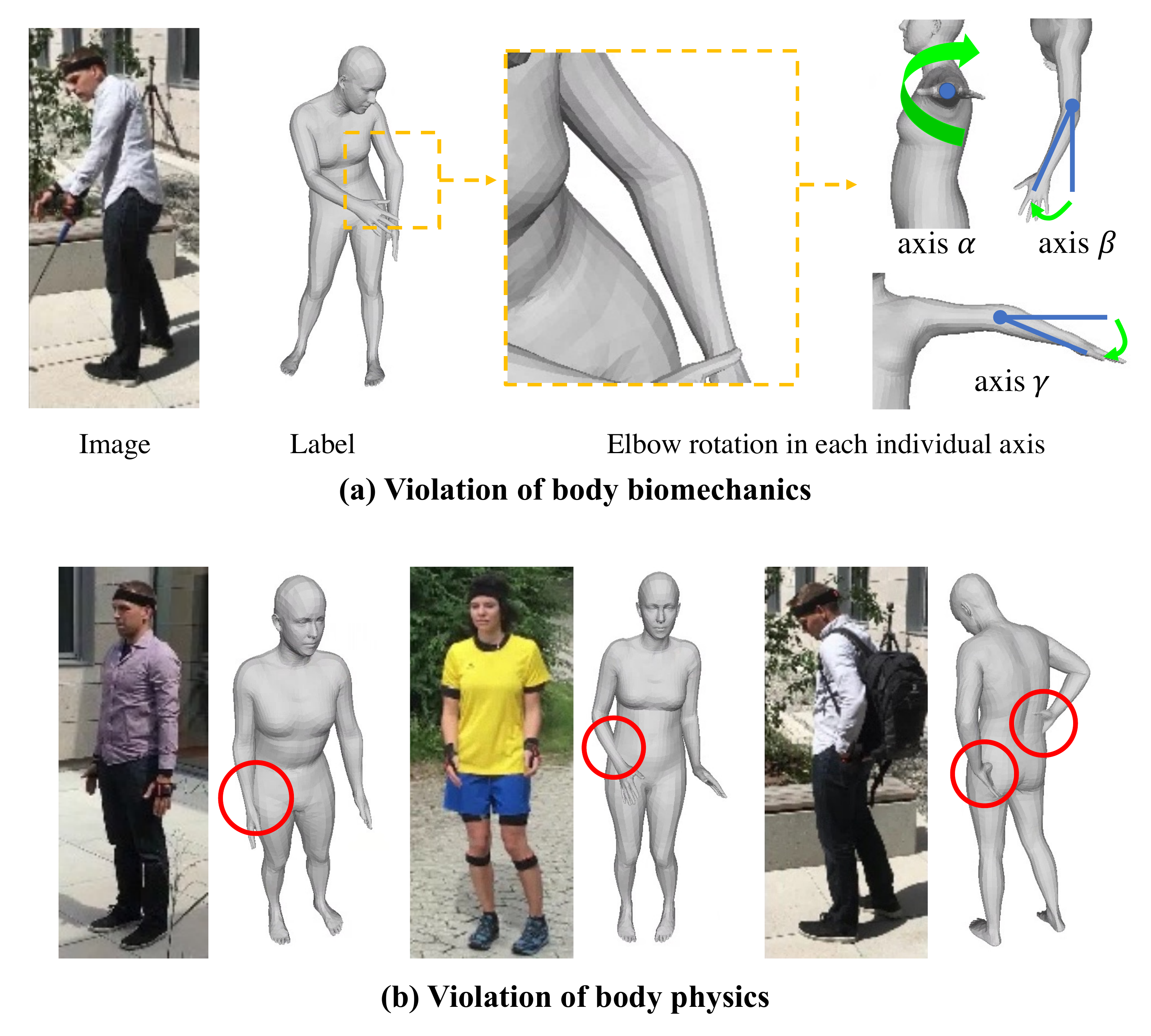}
    \caption{\textbf{Labeling noise presented in existing MoCap data.} The labels in existing MoCap datasets are generated via fitting a parametric body model to a set of sparse 3D markers, which can lead to the violation of (a) body biomechanics; and (b) body physics (the body parts with penetration are marked with red circles) constraints.}
    \label{viobiophy}
\end{figure}

The proposed generic constraints can be used to measure the data noise due to violation of the physical constraints. Example noisy data occurred in existing datasets is shown in Figure~\ref{viobiophy}. In details, in Figure~\ref{viobiophy} (a), the SMPL model annotation is consistent with the original image in general, while the pose label of the left elbow shows infeasible bending. Specifically, for this data sample, $\alpha$, $\beta$, and $\gamma$ at left elbow is -6.4, -24.2, and -18.2 degrees, respectively. While the corresponding valid angle ranges are $(-180,90)$, $(-166,0)$, and $(0,0)$. The rotation defined by $\gamma$ clearly violates the biomechanic constraint and leads to an unrealistic configuration at the left elbow. Furthermore, example violations of body physics are visualized in Figure~\ref{viobiophy}(b). Similarly, the overall configuration given by the label is consistent with the original image but has penetration between body parts, including the unrealistic contact between body hand and torso. These violations of body biomechanics and physics mainly stem from the label generation process, where the labels are generated by fitting to only a set of sparse 3D markers \cite{loper2014mosh,vonMarcard2018}. Although the annotations are generally aligned with the corresponding image data, the violation of the hard generic body constraints leads to unrealistic 3D configuration. We propose to impose the generic body constraints to ensure more physically plausible 3D reconstruction and avoid being affected by the labeling noise.

\end{document}